\documentclass[10pt,twocolumn,letterpaper]{article}

\usepackage[pagenumbers]{cvpr} 

\usepackage{graphicx}
\usepackage{amsmath}
\usepackage{amssymb}
\usepackage{booktabs}
\usepackage{algorithm2e}
\usepackage{float}
\usepackage[pagebackref,breaklinks,colorlinks]{hyperref}

\usepackage[capitalize]{cleveref}
\crefname{section}{Sec.}{Secs.}
\Crefname{section}{Section}{Sections}
\Crefname{table}{Table}{Tables}
\crefname{table}{Tab.}{Tabs.}
\usepackage{booktabs}
\usepackage{multirow}
\usepackage{graphicx}
\def\cvprPaperID{*****} 
\def\confName{CVPR}
\def\confYear{2022}

\begin{document}

\title{CauDR: A Causality-inspired Domain Generalization Framework for Fundus-based Diabetic Retinopathy Grading}

\author{Hao Wei$^1$, Peilun Shi$^1$, Juzheng Miao$^1$, Minqing Zhang$^1$, Guitao Bai$^2$,\\ Jianing Qiu$^1$, Furui Liu$^3$, Wu Yuan$^1$\thanks{Corresponding author: wyuan@cuhk.edu.hk (Wu Yuan).}\\
$^1$The Chinese University of Hong Kong, $^2$Zigong First People's Hospital, $^3$Zhejiang Lab \\  
}

\maketitle

\begin{abstract}
Diabetic retinopathy (DR) is the most common diabetic complication, which usually leads to retinal damage, vision loss, and even blindness. A computer-aided DR grading system has a significant impact on helping ophthalmologists with rapid screening and diagnosis. Recent advances in fundus photography have precipitated the development of novel retinal imaging cameras and their subsequent implementation in clinical practice. However, most deep learning-based algorithms for DR grading demonstrate limited generalization across domains. This inferior performance stems from variance in imaging protocols and devices inducing domain shifts. We posit that declining model performance between domains arises from learning spurious correlations in the data. Incorporating do-operations from causality analysis into model architectures may mitigate this issue and improve generalizability. Specifically, a novel universal structural causal model (SCM) was proposed to analyze spurious correlations in fundus imaging. Building on this, a causality-inspired diabetic retinopathy grading framework named CauDR was developed to eliminate spurious correlations and achieve more generalizable DR diagnostics. Furthermore, existing datasets were reorganized into 4DR benchmark for DG scenario. Results demonstrate the effectiveness and the state-of-the-art (SOTA) performance of CauDR.

\end{abstract}

\section{introduction}\label{sec:introduction}

\begin{figure}[h]
\label{fig_aim}
\centering
\includegraphics[width=0.5\textwidth]{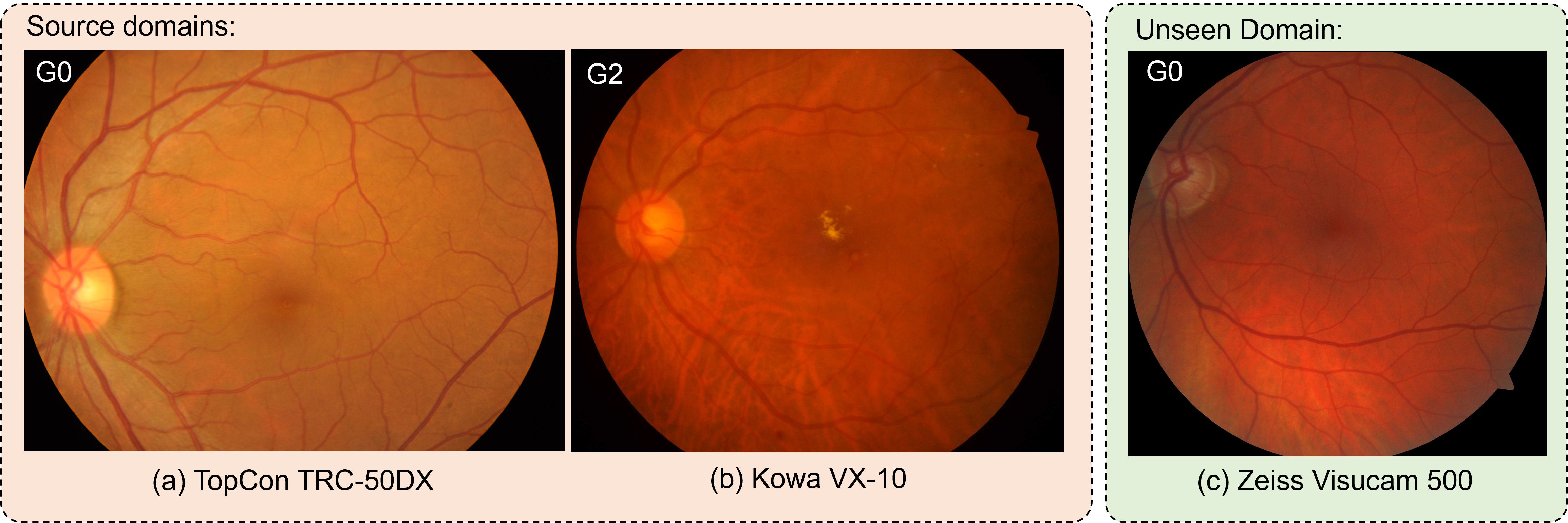}
\caption{Three example fundus images are captured from three imaging devices (a), (b) and (c),
where $G0$ to $G4$ denote the severity grades of DR from healthy to malignant. Once trained on the datasets (source domains) captured from (a) and (b), the model may spuriously correlate the saturation levels of images with DR grades due to the appearance discrepancy induced by imaging devices, i.e., the domain shifts between datasets from (a) and (b). In this case, the trained model may grade image (c) from unseen domain as $G2$ instead of $G0$ based on the saturation levels or the spurious correlations.}. 
\label{fig_confounder}
\vspace{-2.0em}
\end{figure}

{D}{iabetic} retinopathy (DR), the main complication of diabetes, leads to the visual impairment of patients through gradually damaging the retinal blood vessels and diminishing the retinal fundus light-sensitive inner coating~\cite{wang2018diabetic}. In 2022, more than 536.6 million people suffered from diabetes, and more patients are projected in the future~\cite{dridf}. 
If left untreated, DR would affect the patient's vision and cause severe visual impairment, even blindness. Thus, early screening and diagnosis are critical for the effective prevention and treatment of DR. Fundus photography is the most commonly used medical imaging technique to examine the typical pathologies of DR, including hemorrhages, exudates, microaneurysms, and retinal neovascularization~\cite{lee2005computer}. The development of optics and semiconductor technology has promoted the iterative update of cameras rapidly~\cite{liu2022ultrahigh,yao2022developing,ortolano2023quantum,huang2023raman}. Fundus images from innovated cameras typically exhibit various representations~\cite{asaoka2019twofundus}, while the pathological manifestations therein are often minuscule, requiring careful inspections with significant efforts of ophthalmologists. 
Accompanied the number of patients continues to escalate and the burden on the healthcare system becomes increasingly heavy, advanced technologies in imaging and analysis have been developed to facilitate the early detection and optimal management of DR patients. Therefore, a generalized method based on fundus imaging can improve the efficacy of the current time-consuming and labor-intensive DR screening and diagnosis~\cite{das2021recently}.


Recently, deep learning has been successful in tackling various complex computer vision tasks, which attracts researchers to develop deep learning-based DR grading methods~\cite{deng2009imagenet,das2022critical}. Generally, deep learning models are trained to learn the correlations between input and target labels. However, such correlations usually do not imply causality due to the existence of confounder\cite{neuberg2003causality}. As shown in Fig.\ref{fig_confounder}, a DR grading model trained on source domains from imaging devices (a) and (b) may recognize the fundus image from (c) as $G2$ instead of $G0$ based on its high-saturation appearance, causing spurious correlations with DR grades.
In this case, the imaging process of different devices can be regarded as the confounder\cite{neuberg2003causality}, statistically (or spuriously) but not causally correlates with two grades, i.e., $G0$ and $G2$. 
These spurious correlations often degrade the performance on unseen domains due to the lack of these confounders. The discrepancies between the source and unseen domains are termed as domain shifts[9]. To mitigate this issue, 
domain generalization (DG)\cite{DGsurvey} is introduced to train a robust model that can perform well on the data sampled from unseen (target) domains.

During training, DG methods only leverage images from source domains, without the accessibility to target domains. In related publications, most works aim to learn invariant features across domains using meta-learning frameworks\cite{li2018learning}, data augmentations\cite{zhou2021mixstyle}, or new regularization terms\cite{arjovsky2019invariant}. From the perspective of causality\cite{neuberg2003causality}, learning invariant features is equivalent to perform cutting off or so called do-operation on spurious correlations. This usually requires interventions, such as fixing the variable of objects and varying other irrelevant variables\cite{ouyang2022causality}. For example, a straightforward intervention is to take more fundus images with all possible imaging devices for a fixed cohort of patients to amplify source domains. However, this is an unrealistic and impractical way in light of ethical issues and privacy concerns. 

In this work, we propose a causality-inspired framework, i.e., CauDR, to implement virtual interventions on the fundus images to cut off spurious correlations between imaging devices and DR grades. Specifically, to analyze the involved variables and their correlations in fundus imaging, we propose a novel structural causal model (SCM)\cite{neuberg2003causality} to reveal the possible spurious correlations. Then, we identify the spuriously correlated variables in frequency space through discrete cosine transform (DCT)\cite{li2018learning}. Finally, the do-operation is implemented by exchanging the domain variant frequencies while keeping domain invariant ones. In summary, we make the following contributions:

\begin{itemize}
    \item From a causal view, we investigate the DG paradigm for the DR grading task and then propose a novel and effective causality-inspired framework (CauDR) to achieve generalizable performance on the unseen domain.
    \item We propose a novel SCM to help us analyze the involved variables in the imaging process and their correlations. This SCM can also lay the groundwork for the development of new equipment for fundus photography.
    \item To carry out the virtual intervention, we propose 1) a spectrum decomposition step to convert the fundus image into frequency space to simplify the image processing; 2) a novel frequency channel identification strategy to identify spuriously correlated variables; 3) an exchanging-based do-operation to cut off spurious correlations. These three important steps are the cores of our proposed framework.
    \item To evaluate the generalizable performance, we re-organize the existing four DR grading datasets based on imaging devices to build a DR grading benchmark, called 4DR, in the DG scenario, offering opportunities for future works to explore new DG methods.
    \item The proposed CauDR framework attains state-of-the-art performance on the 4DR benchmark, surpassing other domain generalization baseline methods developed for both natural and medical image analysis. Moreover, the in-depth examination of causality and its integration into DR grading conducted in this work may spur the research community to devote greater attention to domain generalization and causality-based approaches. 
\end{itemize}

\section{Related works}\label{sec:related}

\subsection{Grading of Diabetic Retinopathy}
The widely used disease severity classification system for DR \cite{wilkinson2003proposed} includes five grades: normal, mild, moderate, severe non-proliferative, and proliferative, corresponding to $G0$ to $G4$, respectively. Early works on automated DR grading rely on hand-crafted features and utilize machine learning algorithms to classify these features\cite{roychowdhury2013dream}. Recently, deep learning-based DR grading models have been proposed, most of which are based on off-the-shelf models developed for natural image processing or their variants of more effective network structure designs. For example, Gulshan et al.\cite{gulshan2016development} trained an Inception-v3 network on fundus images for the DR grading task. Inspired by this idea, Alantary et al.\cite{al2021multi} introduced a multi-scale network to learn more detailed features. To encourage models to focus on informative regions, attention mechanisms were considered and incorporated into DR grading methods. Xiao et.al\cite{xiao2021se} combined an enhanced Inception module with a squeeze-and-excitation (SE) module and achieved an improved DR grading performance. Later, HA-Net\cite{shaik2022hinge} was proposed to utilize multiple attention stages to learn global features of DR. 
However, the common assumption of independent and identically distributed (I.I.D.) variables among training and testing data in these methods is hard to hold in real-world scenarios\cite{atwany2022drgen}, leading to a degraded performance on unseen data, which compromises the potential clinic values.

\subsection{Domain Generalization}

DG aims to learn invariant features across multiple source domains for a well-generalizable capability on unseen target domains\cite{DG2011,DGsurvey}. 
To achieve this aim, some works assumed exact formats of discrepancies between domains and then designed strategies to reduce these differences. For example, 
Xu et.al \cite{xu2020adversarial} treated image appearances as the main discrepancies between different domains and then proposed domain MixUp to linearly interpolates image-label pairs between domains to reduce the possible spurious correlations between domain and appearance. Later, Nam et.al\cite{nam2021reducing} proposed to disentangle image style and contents to reduce style bias among domains, assuming image styles are sensitive to domain shifts. 

In addition, encouraging a model to learn domain-invariant representations across multiple domains is another strategy for handling DG problems. In this regard, Sun et.al\cite{sun2016deep} proposed a new loss, termed CORAL loss, to minimize discrepancies between the feature covariance across multiple domains in the training set, which explicitly aligns the learned feature distributions to learn domain-invariant representations. Based on a similar strategy, Li et.al\cite{li2018domain} designed a novel framework to match learned feature distributions across source domains by minimizing the Maximum Mean Discrepancy (MMD) and then aligned the matched representations to a prior distribution by adversarial learning to learn universal representations. In addition to the above-mentioned methods taht align features, several recent works have explored aligning gradients between domains to constrain the model's learning. For example, Shi et.al\cite{shi2021gradient} and Rame et.al~\cite{rame2022fishr} incorporated new measures to align inter-domain gradients during the training process, encouraging the model to learn invariant features among domains. 

On the other hand, DG has attracted increasing attention in the field of medical image analysis, where domain shifts are often related to variations in clinical centers, imaging protocols, and imaging devices\cite{DGqidou} \cite{qidou2019domain}.
Li et al.\cite{li2020DGnips} proposed to learn the invariant features by restricting the distribution of encoded features to follow a predefined Gaussian distribution, while Wang et al.~\cite{wang2020dofe} developed the DoFE framework for generalizable fundus image segmentation by enriching image features with domain prior knowledge learned from multiple source domains. Based on Fishr~\cite{rame2022fishr}, Atwany et al.~\cite{atwany2022drgen} recently adopted the stochastic weight averaging densely (SWAD) technique to find flat minima during DR grading training for better generalization on fundus images from unseen datasets.



\subsection{Causality-inspired DG} 
Causality\cite{neuberg2003causality} is a branch of research that explores the connections between various causes and their corresponding effects, with the primary goal to comprehend the underlying mechanisms and patterns behind the occurrence of events. Theoretically, the relationship of $cause \rightarrow effect$ should be the same across domains. For example, an object is considered as a cat because it has the cat's characteristics, which are consistent in different domains. Therefore, the causality-inspired model is a feasible method to tackle DG problems. By considering causality, Arjovsky et. al\cite{arjovsky2019invariant} 
proposed a new regularization term to constrain the optimal classifier (learn $X \rightarrow Y$) in each domain to be the same by minimizing the invariant risk, called Invariant Risk Minimization (IRM). Similarly, Chevalley et.al \cite{chevalley2022invariant} developed a framework CauIRL to minimize the distributional distance between intervened batches in latent space by using MMD or CORAL technique, encouraging the learning of invariant representations across domains. However, these works mainly focus on natural image processing instead of medical image analysis. More recently, Ouyang et.al\cite{ouyang2022causality} proposed to incorporate causality into data augmentation strategy to synthesize domain-shifted training examples in an organ segmentation task, which extends the distributions of training datasets to cover the potentially unseen data distributions. As for our task, the fundus image grading is relatively more challenging due to the complicated factors associated with DR severity. 


\section{Methodology}\label{sec:method}
This section provides a detailed description of our causality-inspired framework. We first formulate the DG problem, followed by a novel causality-based perspective on the image generation process. The proposed methodology is then elaborated.


\subsection{Problem Formulation}

A training image set $S_{train}$ is composed of image pairs $x \in \mathcal{X}$ and its label $y \in \mathcal{Y}$, which are sampled from a joint distribution $P_{XY}^{train}$. The test image pairs are sampled from $P_{XY}^{test}$ to form the set $S_{test}$.
In a regular learning setting, the model $f$ is trained on $S_{train}$ to learn the mapping from image space $\mathcal{X}$ to label space $\mathcal{Y}$ by minimizing the empirical risk $\mathbb{E}_{(x, y) \in S_{train}}[\mathcal{L}(f(x), y)]$, where $\mathcal{L}$ denotes the loss function used in the training phase. The trained model $f$ is expected to perform well on the test set $S_{test}$ if $P_{XY}^{test} = P_{XY}^{train}$, which is often difficult to be satisfied considering the complexity of real-world scenarios. For instance, fundus imaging in different vendors would produce images of diverse appearances.

Current study extends the regular learning paradigm by considering the existence of mismatching between the joint distributions of $P_{XY}$ in training and testing sets, where different $P_{XY}$ are treated as different domains. More specifically, we characterize the training set $S_{train}$ formed by $m$ domains: $S_{train}=\{D^d=\{(x_i^d, y_i^d)\}_{i=1}^n \sim P_{XY}^d \}_{d=1}^m (P_{XY}^i \neq P_{XY}^j)\}$. The goal of domain generalization is to learn a predictor $f$ on $S_{train}$ that can perform well at the unseen test set $S_{test}=\{D^{m+1} \sim P_{XY}^{m+1} \}, S_{test} \cap S_{train} = \emptyset$. 


Due to the condition of $S_{test} \cap S_{train} = \emptyset$, the domain generalization problem\cite{gulrajani2020domainbed} holds a strong assumption that there exist invariant features or information across multiple domains, which is expected to be learned by the predictor $f$ so that it can generalize well to unseen domains. Extracting features that are invariant from the training set is the key to solve the generalization problem. 

It should be noted that domain generalization differs from domain adaptation on accessing unseen domains, where the images from unseen domain $P_{XY}^{m+1}$ are unavailable during the training process in the former but available in the latter. As shown in Table \ref{table_learning}, we demonstrate different learning paradigms to highlight the characteristics of DG settings\cite{gulrajani2020domainbed}.


\begin{table}\centering
\vspace{-2.0em}
\caption{The settings of different learning paradigms, where $L^d$ and $U^d$ denote the labeled and unlabeled distributions from the domain $d$.
}
\begin{tabular}{l|l|l}%
\hline
Learning Paradigm & Training set & Test set \\
\hline
Supervised Learning & $L^d$ & $U^d$ \\
Semi-Supervised Learning & $L^d, U^d$ & $U^d$ \\
Domain Adaptation & $L^1,...,L^m, U^{m+1}$ & $U^{m+1}$ \\
Domain Generalization & $L^1,...,L^m$ & $U^{m+1}$ \\
\hline
\end{tabular}
\label{table_learning}
\vspace{-2.0em}
\end{table}


\subsection{Structural Causal Model}

To understand and analyze the relationships between involved factors in the fundus imaging process, we propose the SCM shown in Fig.\ref{fig_SCM}(a), including the observed image $X$, the context factor $C$, the imaging device factor $D$, and the image label $Y$, where the direct links between nodes denote causalities, i.e., cause $\rightarrow$ effect. 
During the imaging process, the context factor $C$ and the imaging device factor $D$ are logically combined together to generate an observed image $X$ by the internal underlying mechanism (the yellow dotted box in Fig.\ref{fig_SCM}(a)). To illustrate this mechanism, two mediation nodes, including $F_I$ (the domain-invariant features) and $F_V$ (the domain-variant features), are introduced to logically connect $C$, $D$, and $X$. It is worth noting that this SCM can also be applied to other medical modalities if they have a similar imaging process as the fundus.
Specifically, we detail each connection in SCM as follows:

1) $Y \leftarrow C \rightarrow F_I \rightarrow X $: the context factor $C$, such as the retinal vessels or specific lesions, directly determines the imaged contents in the image $X$ through the mediation of the variable $F_I$, which denotes the feature representations of these contents (logically captured by the device $D$). Ideally, $F_I$ only involves the contents specific to patients, without other irrelevant effects from the device (domain-invariant). $F_I \rightarrow X$ denotes the materialization of features into concrete image contents in $X$ (from feature space to spatial space). $Y \leftarrow C$ means the lesions in the retina determine its image-level label, e.g. DR grades in our task. 

2) $D \rightarrow F_V \rightarrow X$: During the imaging process, the device $D$ inevitably introduces irrelevant contents into the final image $X$. In our assumptions, these contents contain not only noise, such as Gaussian noise and speckle noise, but also device-related imaging biases. For example, fundus images of the same patient acquired using different devices may have different hues and saturation due to different hardware and software settings. In this SCM, $F_V$ represents the features of these irrelevant contents, which are device/domain-variant and independent of the context factor $C$. In addition, $F_V \rightarrow X$ means the materialization process, resulting in noise or color bias in the fundus image.

3) $F_I \rightarrow X \leftarrow F_V$: An image $X$ can be regarded as the combination of domain-invariant features $F_I$ and domain-variant features $F_V$, i.e., $g(F_I, F_V) \rightarrow X$, where $g(\cdot)$ denotes the underlying combination and materialization process to generate the observed image $X$. The path  $F_I \leftarrow D \rightarrow F_V$ belongs to the fork structure in causality\cite{pearl2016causal}, where $F_I$ should be is dependent on $F_V$ and $D$ is the confounder.
This dependent relationship brings the spurious correlation between two variables, compromising the generalization ability of the trained model on the images from another unseen imaging device $D$. It should be noted that the underlying mechanism of $F_I \rightarrow X \leftarrow F_V$ in an imaging device is complex and even unknown. The aforementioned process only provides the direction of information flow instead of detailed mechanisms. 

4) $X \rightarrow Y \leftarrow F_I$: Our classification task aims to train a model to learn a function mapping from an image $X$ and its image-level label $Y$: $f=P(Y|X), X \rightarrow Y$. Considering the irrelevant contents (introduced by $F_V$) in $X$, the ideal model should only take $F_I$ as input: $f^*=P(Y|F_I), F_I \rightarrow Y$, in order to gain better generalization ability than that of $f$. Our method aims to reduce the gap between $f$ and $f^*$ to improve generalization ability through the do-operation.

In our study, we assume $C$ and $D$ are inaccessible: ophthalmologists usually choose different imaging devices based on their availability, which means there are no correlations between device $D$ and patients, corresponding to the context factor $C$ by being the captured retina. Therefore, our method implements virtual interventions through do-operation to cut off the path $D \rightarrow F_V$ to make $F_I$ and $F_V$ become independent, reducing the influence induced by spurious correlations between $F_I$ and $F_V$. 
The increased independence between $F_I$ and $F_V$ improves the generalization capability of the model has been proved as well.
In the following section, we describe the implementations of virtual interventions, do-operations, and the structure of the proposed model.


\begin{figure*}[htbp]
\centering
\label{fig_SCM}
\includegraphics[width=1\textwidth]{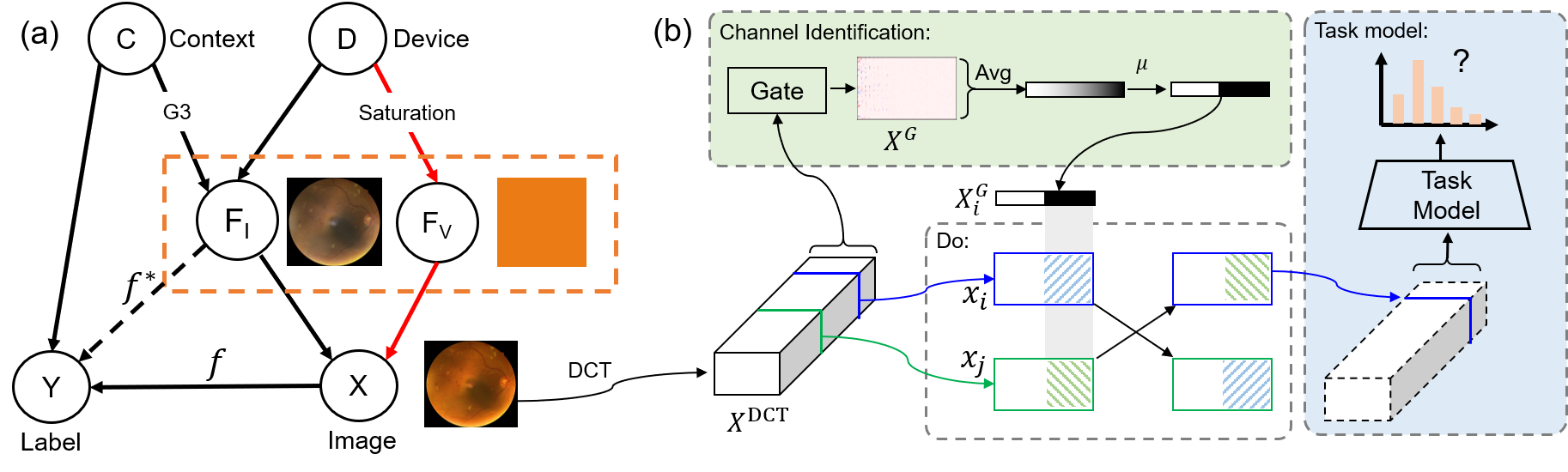}
\caption{(a) The illustration of the proposed structural causal model (SCM) to analyze the involved factors in the fundus image generation process, including observed image $X$, the context factor $C$, the imaging device factor $D$ and also the corresponding image label $Y$. A fundus image ($X$) is captured from a patient ($C$) on an imaging device ($D$), where the yellow dotted box indicates its imaging mechanism. 
During imaging process, different devices $D$ would introduce unknown factors, such as diverse saturation, to affect patients' resulting fundus images $X$. (b) The schematic diagram of our proposed CauDR framework consists of a gate module, a frequency channel identification module, a do-operation step, and a task model. The samples $x_i$ and $x_j$ are presented as rectangles of channel width and height of $H*W$. In the right part of each sample, the rectangle with oblique lines denotes the channels of $F_V$, corresponding to the black region in $\hat{X^G_i}$.} 
\label{fig_SCM}
\vspace{-2.0em}
\end{figure*}

\subsection{Virtual Interventions}

To cut off spurious correlations between domains, an intuitive but impractical strategy is to take fundus images of patients using all possible imaging devices. However, it is impractical . Fortunately, virtual interventions through do-operation can also achieve similar effects. The deep learning-based model in the regular training paradigm is trained to learn $P(Y|X)=P(Y|X=g(F_I, F_V))$. After introducing virtual interventions, our goal becomes to learn a new classifier: $P(Y|do(X)))=P(Y|g(F_I, do(F_V)))$, where $do(\cdot)$ operator is the mathematical representation of intervention. 
Specifically, the do-operation can be implemented by stratifying the confounder, i.e., device factor $D$:

\begin{footnotesize}
\begin{equation}\label{eq1}
\setlength{\abovedisplayskip}{0.5pt}
\setlength{\belowdisplayskip}{0.5pt}
    P(Y|g(F_I, do(F_V)))= \sum_i^{|D|}P(Y|g(F_I, F_{V_i}))P(F_{V_i}|D_i)P(D_i)
\end{equation}
\end{footnotesize}

where $|D|=4$ denotes the number of all available devices. Considering the dependent relationship between $F_V$ and $D$: $D \rightarrow F_V$, we have: $P(F_{V_i}|D_i)=1$. Furthermore,
we adopt the Normalized Weighted Geometric Mean \cite{xu2015show} to move the outer sum $\sum_i P(\cdot)$ into the inner to achieve the batch approximation of Eq.(\ref{eq1}):

\begin{footnotesize}
\begin{equation}\label{eq2}
\setlength{\abovedisplayskip}{0.5pt}
\setlength{\belowdisplayskip}{0.5pt}
    \sum_i^{|D|}P(Y|g(F_I, F_{V_i}))P(D_i) \approx P(Y|\sum_i g(F_I, F_{V_i})P(D_i))
\end{equation}
\end{footnotesize}

However, it is still hard to calculate this equation due to the unknown $g(\cdot)$ function. If we treat this process as a feature fusion procedure, it usually can be implemented as $cat(\cdot)$\cite{he2016resnet}. 
In this study, we adopt the implementation of $cat(\cdot)$ to simplify the complex imaging process and also for better computability. Based on this simplification, we can only change the $F_V$ part for any input image $X$ to achieve Eq.\eqref{eq2} (detailed in subsection E), if we can split the image $X$ into $F_I$ (the domain-invariant features) and $F_V$ (the domain-variant features) parts (detailed in subsection D). 

\subsection{Spectrum Decomposition}
Splitting a fundus image into $F_I$ and $F_V$ is a vital step in our task. For example, considering an image $X$ of a cat on lawn, $X_{fore}$ and $X_{back}$ are used to represent the pixel set of the foreground (cat) and background (lawn), corresponding to $F_I$ and $F_V$ in the high-dimensional feature space, respectively. Then, an ideal classifier $P(Y|X)=P(Y|X_{fore}, X_{back})$ is equivalent to $P(Y|X_{fore})$ considering $X_{back}$ is independent of its label $Y$. Theoretically, we should use $X_{fore}$ only for model training to directly cut off the spurious correlations and achieve an optimal performance. Based on this strategy, Wei et al.~\cite{qin2021causal} proposed a classification framework by first segmenting the foreground $X_{fore}$ through a segmentation network and then taking current $X_{fore}$ and randomized $X_{back}$ as the input to improve the generalization performance on ImageNet\cite{deng2009imagenet}.  

However, it is difficult to figure out which pixel belongs to $X_{fore}$ or $X_{back}$ in a fundus image $X$ because the DR severity and grades are jointly determined by multiple factors~\cite{lee2005computer}. Xu et al. \cite{xu2020learning} found that an image can be decomposed into informative and non-informative parts relative to subsequent vision tasks, where the former can be considered as the foreground $X_{fore}$ and the latter belongs to $X_{back}$. Inspired by this work, we introduce a spectrum decomposition step to convert a fundus image into its frequency space, where the complex task of splitting fore and background can be implemented as a simple task to identify informative and non-informative frequencies. 

For an image $x \subset R^{H \times W \times 3 }$, Discrete Cosine Transform (DCT) is first used to convert it from the spatial image space into the frequency space. Then, band-pass filters are employed to decompose the frequency signals into 64 channels: 
\begin{equation}\label{eq3}
    X^{DCT} = cat(\{BF(DCT(X), i, j)\}) \subset R^{H \times W \times 3 \times 64}
\end{equation}
where $BF(,i, j), i= \{0,1,2\}, j \subset \{0,...,63\}$ denotes the $j^{-th}$ band-pass filter executed on $i^{-th}$ channel of frequency signals. 
In practice, the DCT coefficients in the JPEG format images are treated as the frequency representations, where different blocks of coefficients approximately implement the filtering process, 
i.e., $X^{DCT} \subset R^{ \frac{H}{8} \times \frac{W}{8} \times 3 \times 64} $ for the input image $x \subset R^{H \times W \times 3}$. Then, the frequency channels can be decomposed into the subsets of $F_I$ and $F_V$. In this case, $iDCT(g(F_I, F_V))$ denotes $F_I \rightarrow X \leftarrow F_V$, where $g(\cdot) \triangleq cat(\cdot)$. 
The method to identify frequencies associated with $F_I$ and $F_V$ is detailed in the following section.



\subsection{Frequency Channel Identification}\label{channel_identify}

Based on our hypothesis, different channels (frequencies) in $X^{DCT}$ carry diverse information for the downstream tasks. To split them into two parts, the informative (salient channels, $F_I$) and non-informative (trivial ones, $F_V$) parts, we introduce a channel attention-based gate module by optimizing an external regularization in the training objectives. 




For the $i^{-th}$ sample, $X_i$, in a batched input, this gate module estimates whether its channels belong to $F_I$ or $F_V$ by reaching the trade-off between fewer channels in $F_I$ and higher performance. During implementation, there are four strategies to achieve channel identification: original estimation from the gate module (OE), batched original estimation (BOE) with continuous values, batched average estimation (BAE) with threshold, and the pre-defined indexes (FE) belonging to $F_I$ in natural images \cite{xu2020learning}, as shown in Fig.\ref{fig_do_ablation}. Based on the ablation study about these four choices, we adopt the average estimation in a batch as the final estimation of a single sample:

\begin{figure}[h!]
\centering
\includegraphics[width=0.4\textwidth]{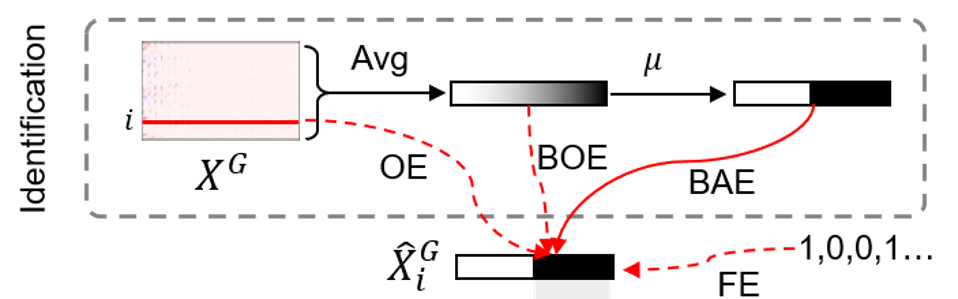}
\caption{Schematic illustration of four channel identification (red arrows) strategies, where the solid arrow denotes the final adopted one. }
\label{fig_do_ablation}
\vspace{-1.0em}
\end{figure}

\begin{equation}\label{eq4}
    \hat{X^G_i} = \mathbb I (\frac{1}{B} \sum_{i=1}^B X^G_i > \mu)
\end{equation}
where $\mathbb I$ is an indicator function to output $1$ if the input condition is satisfied. For the $i^{-th}$ sample in batch inputs (batch size is B), $\hat{X^G_i} \subset  R^{1 \times 192 \times 1 \times 1} $ denotes its batch averaged channel estimation.
$\mu$ is a threshold hyper-parameter to control the estimation sensitivity, where a larger $\mu$ value leads to a smaller set of $F_I$. The selection of $\mu$ is discussed in the Ablation Analysis section.

\subsection{Do-operation}\label{do-operation}

\begin{figure}[h!]
\centering
\includegraphics[width=0.4\textwidth]{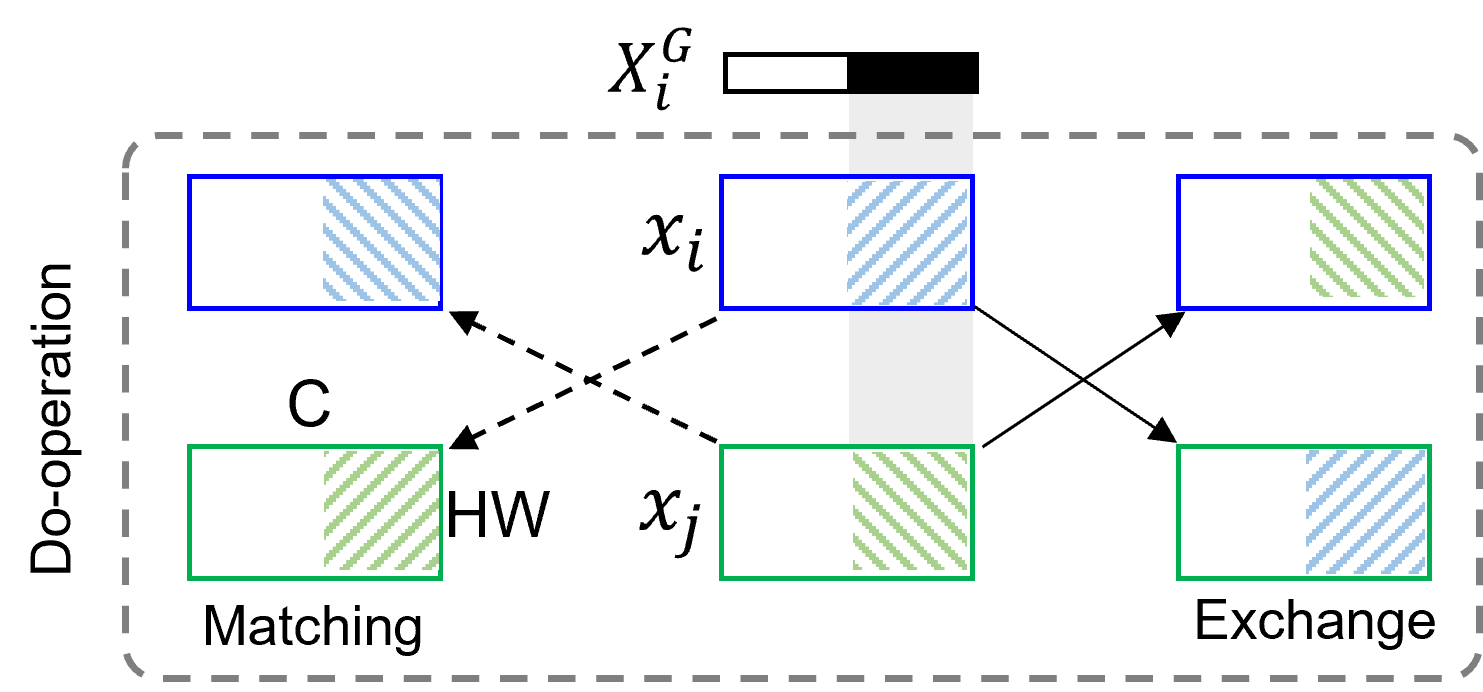}
\caption{The illustrations of two strategies to implement do-operation, where the left and right denote the matching and exchanging-based operations, respectively. $X_i^G$ means the estimation of channel identification in eq.\ref{eq4}}
\label{fig_do_operation}
\vspace{-1.0em}
\end{figure}

After the identification of frequency channels, we now consider the implementation of the do-operation in Eq.\ref{eq2}, i.e., iterating different $F_V$ while keeping $F_I$ unchanged. Generally, there are two strategies, as shown in Fig. \ref{fig_do_operation}: 1) directly exchange the channels of $F_V$ between two arbitrary input samples, and 2) exchange the statistical characteristics of frequency channels between two input samples, such as the mean and standard deviation. We hypothesize that the co-occurrence of $F_V$ and $F_I$ in one image will lead to spurious correlations during training. Therefore, the first method is able to cut off this co-occurrence relation directly by randomizing $F_V$, which is also validated in subsequent experiments.

Based on the experiments, our framework adopts the exchaning-based strategy to implement do-operation during the training phase, as shown in the right part of Fig.\ref{fig_do_operation}. For each input, $x_i$ and $x_j$, and corresponding frequency representation $x^{DCT}_i$ and $x^{DCT}_j$, the channels belonging to $F_V$ (predicted as 0 in the gate module) are exchanged between them to form new samples, which are inputs of the subsequent networks. By constantly and randomly sampling inputs, spurious correlations between devices $D$ and fundus images $X$ are disordered to cut off $D \rightarrow F_V$, 

\subsection{Network Structures}
\begin{figure}[h!]
\centering
\includegraphics[width=0.4\textwidth]{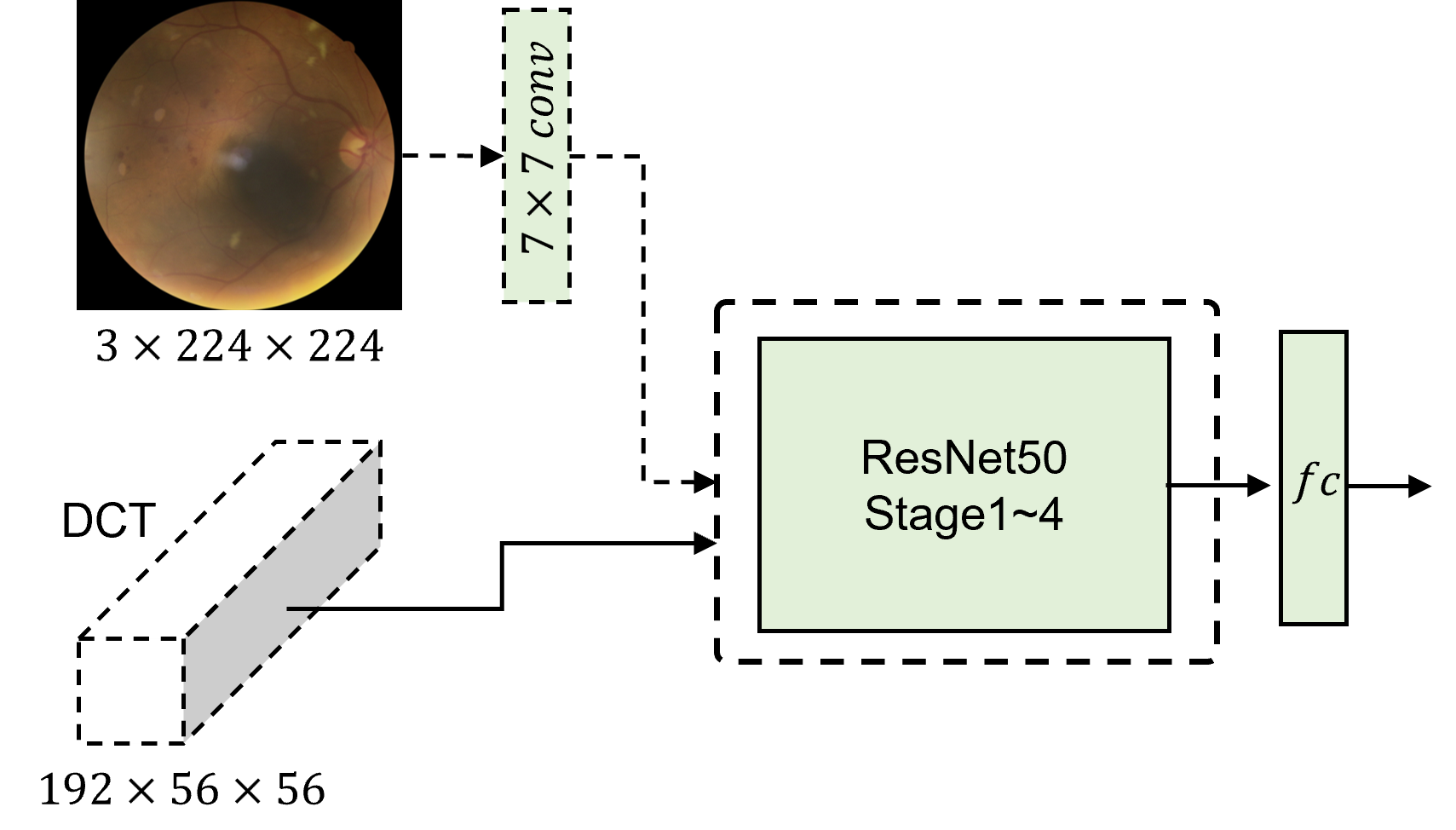}
\caption{The structure of our task model, where the solid arrow denotes our adopted pipeline.}
\label{fig_network}
\vspace{-1.0em}
\end{figure}

The structure of the task model in our framework is shown in Fig.\ref{fig_network}, which is based on the ResNet50. In our implementation, we remove the input layers ($7 \times 7$ convolution), the dashed arrow, to adapt to the DCT coefficients after do-operation. 
During training, the gate module is constrained by the regularization term: $L_{gate} := \sum_{j=1}^{C=192}X^G(j)$, where $j$ denotes the $j^{-th}$ channel in $X^G$. This term would select fewer input channels for $F_I$ (the remaining channels belong to $F_V$). There is an adversarial training schema between $L_{gate}$ and classification performance: lower $L_{gate}$ means fewer $F_I$, which would degrade the accuracy ($F_I$ is more informative than $F_V$). 

In addition, do-operation requires the combination of both $F_I$ and $F_V$ channels. During training, however, the first $1 \times 1$ convolutional layer may block the trivial $F_V$ channels by zero weights. To avoid this situation, we introduce a new regulation term to constrain it to select at least half of the channels: $L_{fuse} = max(0.5 - L_1, 0)$, where $L_{1}$ denotes the $L_{1}$ normalization of the first layer.

Finally, the standard cross-entropy loss $L_{CE}$ is used to optimize the classification performance. By incorporating the above-mentioned regularization terms, the objective function of our model is shown below:
\begin{equation}\label{eq_loss}
    L = L_{CE} + \lambda L_{gate} + L_{fuse}
\end{equation}
where $\lambda$ indicates the relative weight of the $L_{gate}$ and is empirically set to 0.18.
\section{Results}\label{sec:results}

\subsection{Dataset and DR Grading}

In order to validate the effectiveness of the proposed framework, we collected 4 public datasets containing subjects from India~\cite{IDRID}, China~\cite{deepdrid,lin2020sustech} and Paraguay~\cite{DR2021dataset}. Then, we curated them into 4 domains based on imaging devices to build a DR grading benchmark, termed 4DR. The 4DR benchmark contains a total of 5082 images and five DR grades ($G0$ to $G4$), as shown in Table. \ref{tab_data}. Other classes irrelevant to DR grades $G0$ to $G4$ in each dataset were discarded to focus on the current task of DR grading.

\begin{table*}[]
\centering
\caption{4DR benchmark for DR grading in the domain generalization setting and their class distribution as percentages of the total number of images in five grades.}
\begin{tabular}{c|c|c|l|l|l|l|l|l|c}%
\hline
Dataset & Domain No.& Device Types & G0 & G1 & G2 & G3 & G4 & Total & Origin\\
\hline
IDRiD\cite{IDRID}  & Domain 1 & Kowa VX-10 & 168 & 25 & 157 & 86 & 58 & 494 & India\\
DeepDRiD\cite{deepdrid}  & Domain 2 & Topcon TRC-NW300 & 918 & 222 & 396 & 352 & 112 & 2000 & China\\
Sustech-SYSU~\cite{lin2020sustech}  & Domain 3 & Topcon TRC-50DX & 631 & 24 & 365 & 73 & 58 & 1151 & China\\
DR2021 Version 0.3\cite{DR2021dataset} & Domain 4 & Zeiss Visucam 500 & 711 & 6 & 110 & 349 & 261 & 1437 & Paraguay\\
\hline
\end{tabular}%
\label{tab_data}
\end{table*}





\subsection{Implementation Details}

The proposed framework was implemented using Pytorch. During training, the pre-trained weight on the ImageNet dataset was first loaded.
Then, we utilized the SGD optimizer to train the model in $5000$ steps with a learning rate of $5e^{-4}$ and weight decay of $1e^{-4}$. The learning rate was decayed by multiplying $0.1$ after reaching $30\%$ and $60\%$ of the total steps. To accelerate the forward process, we adopted the offline DCT transformation by first reading DCT coefficients from the JPEG images and then saving the DCT results into the local disk for fast online loading during training, where the size of the JPEG image and DCT coefficients are $448\times 448 \times 3$ and $56 \times 56 \times 192$, respectively.
Only random horizontal flip is used as the basic data augmentation strategy in the pre-processing step. All experiments were conducted on one NVIDIA RTX 3060 12GB GPU with a batch size of 96. Each experiment was repeated three times by using different random seeds.

In addition, we implemented most of the comparable methods by using the code in the DomainBed\cite{gulrajani2020domainbed} benchmark. For the algorithms not in this benchmark, we utilize their official implementation to ensure fair comparisons.

\subsection{Experimental Results on DR Grading}

\begin{table}[]
\caption{Quantitative comparison with domain generalization methods on the DR grading task, where the results of each domain denote the model was trained on the images from the other three domains. We report the mean performance and standard error after three trials for each experimental setting. The top results are highlighted in \textbf{bold}.}
\setlength\tabcolsep{2pt}
\resizebox{\linewidth}{!}{
\begin{tabular}{@{}lccccc@{}}
\toprule
\multirow{2}{*}{Network} & \multicolumn{5}{c}{Performance}                                                    \\ \cmidrule(l){2-6} 
                         & Domain1        & Domain2        & Domain3        & Domain4        & Average        \\ \midrule
ResNet50~\cite{he2016resnet}                 & $52.86_{(1.69)}$    & $48.56_{(1.82)}$    & $65.18_{(0.97)}$    & $55.07_{(0.95)}$    & 55.42          \\ \midrule

Mixup\cite{xu2020adversarial}                    & $53.53_{(0.15)}$    & $48.92_{(0.42)}$    & $65.76_{(2.81)}$    & $63.74_{(1.72)}$    & 57.99          \\
SagNet\cite{nam2021reducing}                   & $53.62_{(0.86)}$    & $47.56_{(0.99)}$    & $62.80_{(2.19)}$    & $61.62_{(1.48)}$    & 56.40          \\ \midrule
MMD\cite{li2018domain}                      & $56.14_{(0.13)}$    & $49.21_{(2.94)}$    & $65.33_{(2.56)}$    & $60.46_{(1.29)}$    & 57.79          \\
CORAL\cite{sun2016deep}                    & $52.53_{(0.82)}$    & $49.62_{(2.94)}$    & $65.25_{(1.47)}$    & $59.04_{(2.84)}$    & 55.86          \\
IRM \cite{arjovsky2019invariant}                      & $58.16_{(1.35)}$    & $47.12_{(0.67)}$    & $66.09_{(3.95)}$    & $58.52_{(1.56)}$    & 57.47          \\
Fish\cite{shi2021gradient}                     & $54.29_{(1.19)}$    & $44.19_{(2.30)}$    & $67.43_{(1.26)}$    & $59.56_{(1.52)}$    & 56.37          \\
Fishr\cite{rame2022fishr}                    & $57.24_{(2.19)}$    & $47.73_{(2.43)}$    & $67.61_{(0.77)}$    & $56.29_{(1.62)}$    & 57.22          \\
DRGen\cite{atwany2022drgen}                    & $55.97_{(1.23)}$    & $50.00_{(1.18)}$    & $59.43_{(3.63)}$    & $\textbf{67.85}_{(1.78)}$    & 58.31          \\ \midrule
Cau\_MMD\cite{chevalley2022invariant}             & $55.81_{(1.68)}$    & $48.20_{(0.84)}$    & $62.68_{(3.83)}$    & $58.86_{(3.37)}$    & 56.41          \\
Cau\_CORAL\cite{chevalley2022invariant}           & $51.85_{(0.22)}$    & $51.41_{(0.88)}$    & $67.82_{(3.02)}$    & $56.99_{(6.15)}$    & 57.01          \\ \midrule
CauDR(Ours)              & $\textbf{60.52}_{(0.12)}$    & $\textbf{54.17}_{(1.08)}$    & $\textbf{72.67}_{(1.29)}$    & $59.91_{(1.62)}$    & \textbf{61.82} \\ \bottomrule
\end{tabular}}
\label{result_table}\vspace{-2.0em}
\end{table}

\subsubsection{Comparison With Baseline Models}

We trained the commonly used ResNet50~\cite{he2016resnet} on all source domains in a regular learning manner (I.I.D assumption) as the baseline models. As shown in Table \ref{result_table}, our proposed CauDR outperforms the baseline by significant margins (average accuracy increase of $6.4\%$), which clearly demonstrates the effectiveness and better generalization ability of our method.

In addition, the baseline under-performs all the DG-optimized methods, demonstrating the necessity of study in domain generalization issues.

\subsubsection{Comparison with Appearance-based Methods}
In MixUp\cite{xu2020adversarial} and SagNet\cite{nam2021reducing}, authors assumed the image appearance differences result in discrepancies across domains. Therefore, they designed strategies to randomize or remove appearance while keeping the contents fixed. 
As shown in Table \ref{result_table}, both MixUp and SagNet achieve better results than ResNet50.
However, those assumptions are limited as they only consider the image appearance in spatial space. In contrast, our proposed frequency-based operation covers more types of discrepancies and produces performance improvements of $3.83 \%$ and $5.42 \%$ when compared with MixUp and SagNet, respectively.

\subsubsection{Comparison with Representation-based Methods}

Instead of directly manipulating the input data, designing strategies in feature or gradients space to encourage learning domain-invariant representations across domains is also one of the main directions to solve DG problems. We implemented features alignment-based (CORAL\cite{sun2016deep}, MMD\cite{li2018domain}) and gradients alignment-based (Fish\cite{shi2021gradient}, Fishr\cite{rame2022fishr}, DRGen\cite{atwany2022drgen}) methods on our task
to evaluate their performance. As shown in Table \ref{result_table}, MMD achieves $57.79\%$ in average accuracy while CORAL performs poorly but still slightly outperforms ResNet50. 


On the other hand, Fish and Fishr adopted new measures to align gradients across domains to promote the learning of domain-invariant features, while DRGen incorporated the SWAD technique into Fishr to search for the flat minima during training. As shown in the Table \ref{result_table}, Fish, Fishr, and DRGen outperform the ResNet50 but fail to surpass our method. 
This may reveal the challenges to learn invariant representations by enforcing gradient directions on the small and class-imbalanced fundus datasets. 

\subsubsection{Comparison With Causality-based Methods}
Currently, there are only limited studies focused on the causality for the DG problems in image classification tasks, such as DR grading. IRM\cite{arjovsky2019invariant} assumed the underlying causal relationship (observable features $X \rightarrow$ interested variable $Y$) in each domain is constant and then proposed a regularization term to constrain the optimal classifier (learn $X \rightarrow Y$) in each domain to be the same. Similarly, CauIRL\cite{chevalley2022invariant} was developed to minimize the distributional distances (measured by MMD or CORAL, termed Cau\_MMD and Cau\_CORAL, respectively) between batches intervened by the confounder in the latent space to encourage the learning of invariant representations under the pre-defined SCM. As shown in Table \ref{result_table}, the performance of IRM slightly outperforms CauIRLs, i.e. Cau\_MMD and Cau\_CORAL. However, IRM performs worse than other DG methods and its performance is inferior to our method by a considerable margin of $4.35\%$. The complicated invariant and variant features in our task may result in this performance gap, where optimizing the invariance of the classifier alone is not sufficient to learn the domain-invariant features. Likewise, the regularization of encoded features in CauIRLs may not be effective in handling current complicated scenarios. 
In addition, the implementation of intervention is important and the results demonstrate that our exchange-based intervention outperforms the distance constrain-based intervention in CauIRLs. More experiments on the implementations of intervention are detailed in the Ablation Analysis section.



\subsection{Ablation Analysis}


\begin{table}[]
\caption{Ablation study of our used loss terms. "w/o Gate" denotes without $L_{gate}$ term in Eq.\ref{eq_loss} while "w/o Fuse" means the removal of $L_{fuse}$ in Eq.\ref{eq_loss}. The top results are highlighted in \textbf{bold}}
\setlength\tabcolsep{4pt}
\resizebox{\linewidth}{!}{
\begin{tabular}{@{}llllll@{}}
\toprule
Method   & Domain 1 & Domain 2 & Domain 3 & Domain 4 & Average \\ \midrule
w/o Gate & $58.92_{(0.22)}$       & $53.81_{(0.62)}$       & $72.06_{(0.59)}$       & $59.39_{(1.27)}$       & 61.05      \\
w/o Fuse & $58.50_{(0.63)}$       & $54.06_{(0.85)}$       & $\textbf{72.71}_{(0.02)}$       & $58.90_{(0.68)}$       & 61.04      \\
Ours     & $\textbf{60.52}_{(0.12)}$       & $\textbf{54.17}_{(1.08)}$       & $72.67_{(1.29)}$       & $\textbf{59.91}_{(1.62)}$       & \textbf{61.82}      \\ \bottomrule
\end{tabular}}
\label{loss_ablation_table}
\vspace{-2.0em}
\end{table}


\subsubsection{Effectiveness of Loss Terms in Eq.\ref{eq_loss}}
\begin{figure}[h!]
\centering
\includegraphics[width=0.4\textwidth]{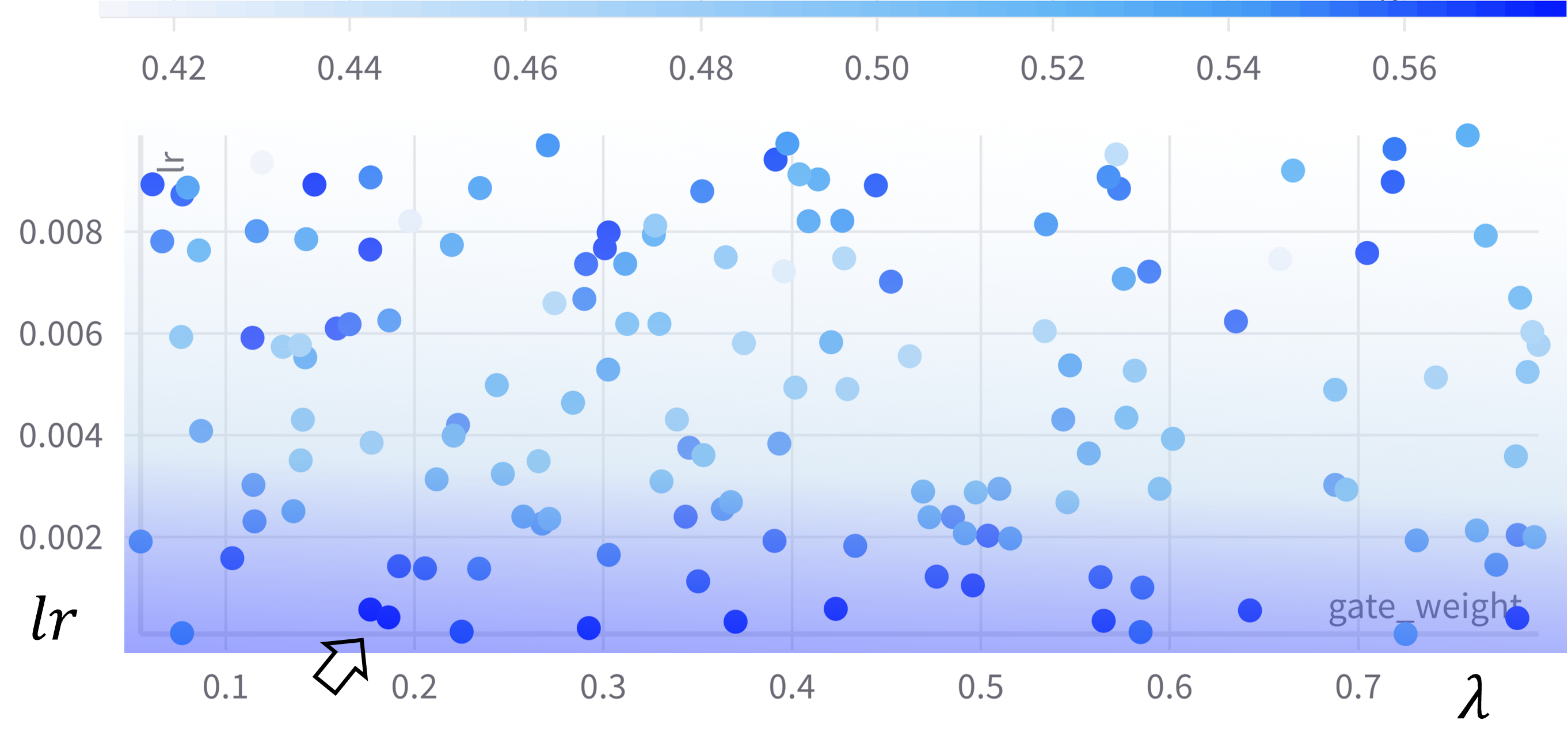}
\caption{Each point denotes a trial in hyper-parameter searching, and its color indicates the performance (the bluer the better), where the horizontal and the vertical axes denote the $\lambda$ in $[0.05, 0.8]$ and $lr$ in $[2.5e^{-5}, 1e^{-2}]$. The black arrow points to the best results at $\lambda=0.18$ and $lr=5e^{-4}$}
\label{fig_hyper}
\end{figure}

Firstly, we utilize the hyper-parameters searching tool in wandb\cite{wandb} to find the optimal $\lambda$ and the learning rate. From Fig.\ref{fig_hyper}, we determine the acceptable parameters: $\lambda = 0.18$ and $lr=5e^{-4}$. Then, we evaluate the effectiveness of gate and fuse regularization terms by removing one of them. The average performance drop (about $0.8\%$) can be observed from Table \ref{loss_ablation_table} after removing the gate or fuse term. Removing the gate loss term may degrade the channel identification accuracy, further reducing the generalization ability of the model. Similarly, the removal of the fuse loss term also reduces the performance gain of the proposed virtual interventions.

\subsubsection{Influence of $\mu$ in Eq.\ref{eq4}}

We further explored the influence of $\mu$ in Eq.\ref{eq4} by separately setting its value to 0.1, 0.2, 0.4, 0.6, 0.8, and 0.9. From the results in Table \ref{mu_ablation_table}, it can be observed that the average accuracy increases first and then decreases along with the increase of $\mu$. Therefore, we chose $\mu =0.4$ in Eq.\ref{eq4} for the optimal performance.

\begin{table}[]
\caption{Quantitative results with different $\mu$ in Eq.\ref{eq4}, where top results are highlighted in \textbf{bold}}
\resizebox{\linewidth}{!}{
\begin{tabular}{@{}llllll@{}}
\toprule
$\mu$   & Domain 1 & Domain 2 & Domain 3 & Domain 4 & Average \\ \midrule
0.1 & $58.42_{(1.02)}$       & $53.85_{(0.85)}$       & $72.57_{(1.61)}$       & $60.03_{(1.71)}$       & 61.22      \\
0.2 & $59.09_{(1.26)}$       & $53.65_{(0.66)}$       & $71.66_{(1.57)}$       & $\textbf{60.12}_{(1.72)}$       & 61.13      \\
0.4 & $\textbf{60.52}_{(0.12)}$       & $\textbf{54.17}_{(1.08)}$       & $72.67_{(1.29)}$       & $59.91_{(1.62)}$       & \textbf{61.82}      \\
0.6 & $58.33_{(1.56)}$       & $53.79_{(1.14)}$       & $\textbf{73.18}_{(1.83)}$       & $59.48_{(1.67)}$       & 61.20      \\
0.8 & $58.50_{(0.32)}$       & $53.21_{(0.97)}$       & $73.04_{(1.91)}$       & $60.12_{(2.17)}$       & 61.21      \\
0.9     & $58.92_{(0.73)}$       & $53.63_{(0.71)}$       & $72.13_{(1.75)}$       & $59.83_{(1.79)}$       & 61.13      \\ \bottomrule
\end{tabular}}
\label{mu_ablation_table}
\vspace{-2.0em}
\end{table}



\subsubsection{Channel Identification and Do-operation}
In section \ref{channel_identify} and \ref{do-operation}, we introduce four strategies to implement the channel identification, as shown in Fig.\ref{fig_do_ablation}, and two ways for the do-operation in Fig.\ref{fig_do_operation}. Therefore, eight methods can be achieved by randomly combining strategies of channel identification and do-operation implementation, as shown in Table \ref{do_ablation_table}.

\begin{table}[]
\caption{Ablation experiments of channel identification and do-operation implementations, where Ex+ and Ma+ denote exchange and match strategies, respectively. top results are highlighted in \textbf{bold}}
\setlength\tabcolsep{4pt}
\resizebox{\linewidth}{!}{
\begin{tabular}{@{}llllll@{}}
\toprule
Do   & Domain 1 & Domain 2 & Domain 3 & Domain 4 & Average \\ \midrule
Ex+OE & $59.09_{(0.67)}$       & $53.21_{(0.81)}$       & $72.86_{(1.04)}$       & $60.41_{(0.98)}$       & 61.39      \\
Ex+BOE & $59.43_{(1.03)}$       & $53.60_{(0.61)}$       & $72.46_{(0.86)}$       & $60.00_{(1.24)}$       & 61.37      \\
Ex+BAE & $\textbf{60.52}_{(0.12)}$       & $\textbf{54.17}_{(1.08)}$       & $72.67_{(1.29)}$       & $59.91_{(1.62)}$       & \textbf{61.82}      \\
Ex+FE & $60.27_{(0.59)}$       & $52.18_{(0.81)}$       & $\textbf{73.01}_{(1.34)}$       & $\textbf{61.07}_{(0.68)}$       & 61.79      \\
\hline
Ma+OE & $56.09_{(0.67)}$       & $54.21_{(0.72)}$       & $71.37_{(1.60)}$       & $53.54_{(2.24)}$       & 58.96      \\
Ma+BOE & $56.31_{(0.87)}$       & $53.96_{(0.40)}$       & $70.76_{(1.59)}$       & $54.61_{(2.23)}$       & 58.91      \\
Ma+BAE & $57.32_{(1.62)}$       & $54.00_{(0.51)}$       & $71.33_{(1.57)}$       & $53.24_{(2.57)}$       & 58.97      \\
Ma+FE & $55.13_{(1.10)}$       & $53.94_{(0.34)}$       & $71.66_{(1.41)}$       & $54.14_{(1.18)}$       & 58.76      \\\bottomrule
\end{tabular}}
\label{do_ablation_table}
\vspace{-1.0em}
\end{table}

From the results, the exchange-based methods outperform the match-based methods by a margin of circa 2.5 $\%$ on the average accuracy, indicating the effectiveness of the exchanging mechanism, although the matching-based variants perform better than other DG-related algorithms.
Furthermore, we found that different methods in either exchange-based or match-based group achieve similar performance, signifying the pre-trained weights in natural images still provide correct estimations on $F_I$ and $F_V$ in fundus images. Eventually, we adopted BAE strategy in our method for its optimal performance.

\subsubsection{Visualizations of Do-operation}

\begin{figure}[h!]
\centering
\includegraphics[width=0.4\textwidth]{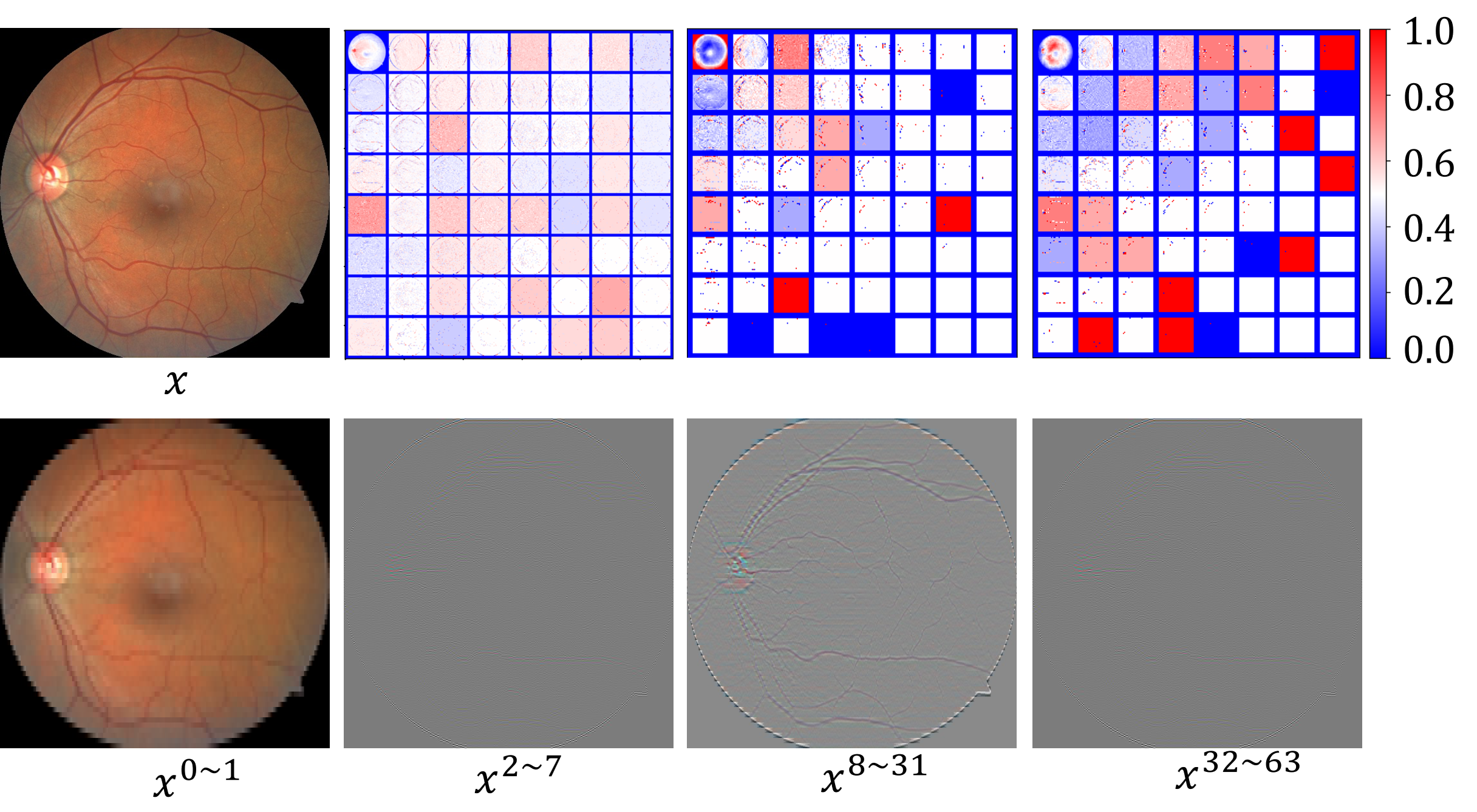}
\caption{Visualizations for DCT coefficients of image $x$ and images reconstructed from different coefficient subsets, where $x ^ {m \sim n}$ denotes the image reconstructed from the $m^{-th}$ to $n^{-th}$ channel in frequency space. Zoom in for best view.}
\label{fig_DCT}
\vspace{-1.0em}
\end{figure}

For an arbitrary image $x$, we first visualize its DCT coefficients ($3 \times 64 = 192$ channels) in Fig.\ref{fig_DCT} with three maps, corresponding to RGB channels in the spatial space, where each map consists of $8 \times 8 $ channels (small images). In each map, the frequency gradually increases from the upper left to the bottom right by a $z$ shape path\cite{xu2020learning}. 

From Fig. \ref{fig_DCT}, we find that the low frequencies usually contain more contour information, such as the vessels and the optic disk. As the frequency increases, more detailed local information becomes available. We also observe this phenomenon in the reconstructed images in the second row by exchanging the subset of frequencies between two images.

\begin{figure}[h!]
\centering
\includegraphics[width=0.4\textwidth]{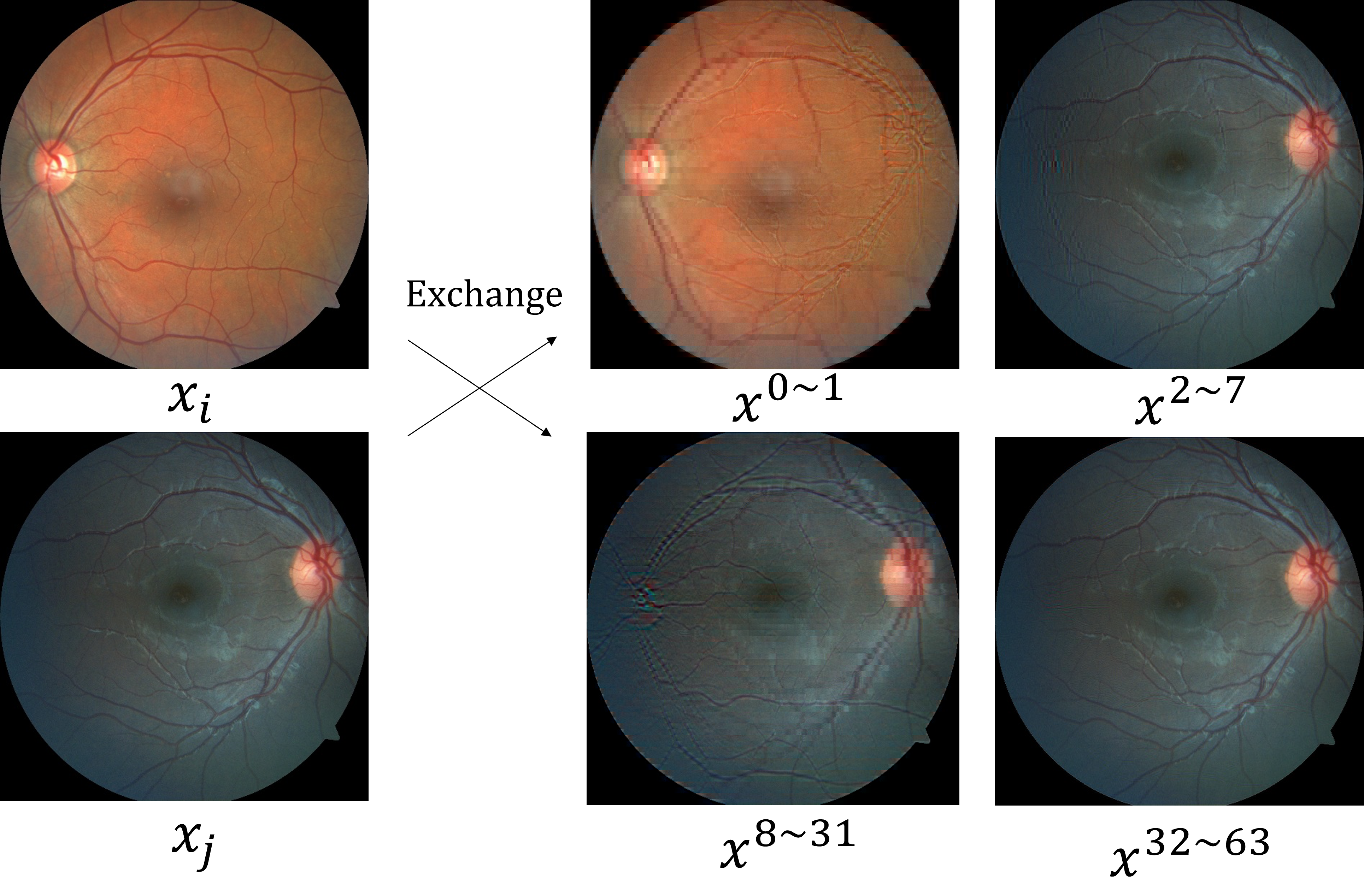}
\caption{The illustration of do-operation between images $x_i$ and $x_j$ by keeping $m^{-th}$ to $n^{-th}$ channels in $x_i$ and exchanging the remaining channels with $x_j$. Zoom in for best view.}
\label{fig_do_img}
\vspace{-1.0em}
\end{figure}

Then, we randomly sample images $x_i$ and $x_j$ to visualize the effects of do-operation: exchanging the subset of frequencies between them. As shown in Fig.\ref{fig_do_img}, the intervented images simultaneously include the contents from two images in different frequencies, where the domain-related information is expected to be disordered by this exchange.

\subsubsection{Visualizations for Feature Representations}
\begin{figure}[h!]
\centering
\includegraphics[width=0.4\textwidth]{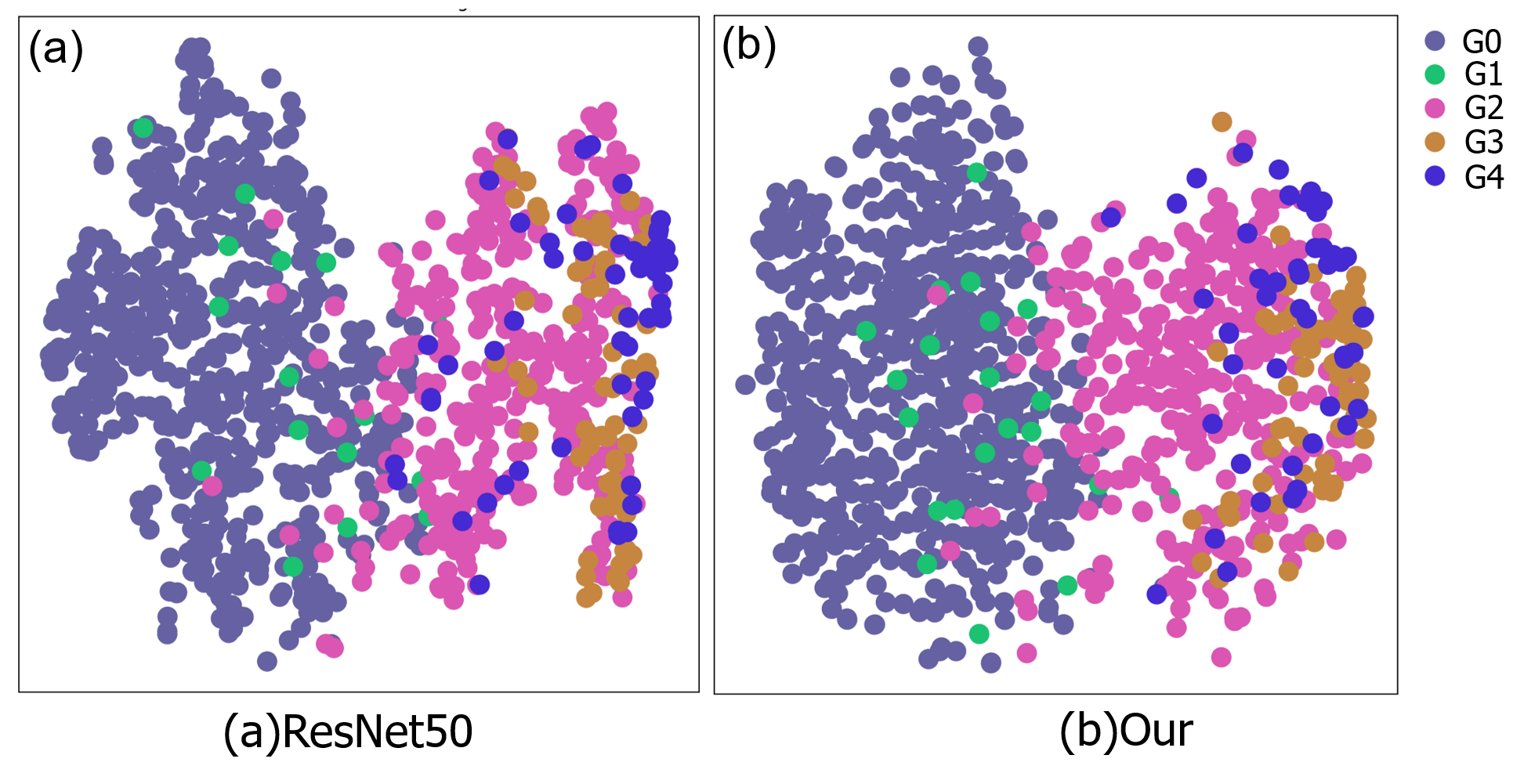}
\caption{The t-SNE visualizations of feature representations of Domain 2 dataset extracted through the baseline ResNet50 (a) and our method (b). Colors denote different DR grades.}
\label{fig_cls_tsne}
\vspace{-2.0em}
\end{figure}

We utilize the t-SNE\cite{van2008visualizing} technique to visualize the learned feature representations extracted from the Domain 2 dataset using the baseline ResNet50 and our method. As shown in Fig.\ref{fig_cls_tsne}, the class representations extracted using our method are more sparsely separated than those extracted using the baseline method. For example, the representations of $G4$ are mixed with those of $G2$ and $G0$ in Fig.\ref{fig_cls_tsne}(a), while this mixing effect is not obvious in Fig.\ref{fig_cls_tsne}(b). This indicates that the features extracted by our method have a clear decision boundary, which is beneficial for the classification task.

\section{Discussion}\label{sec:discuss}


Our proposed method demonstrates better generalizable performance on DR grading compared with other DG methods. Nevertheless, our method still has limitations. First, the limited source domains in the training set hinder the learning of invariant features across domains due to the lack of enough invariability. Collecting more images from different devices would alleviate this issue and further improve the overall generalizable performance on unseen datasets. However, in practice, collecting and annotating more fundus images is troublesome and expensive when considering the involved privacy and ethical matters and the clinical experience of annotators. One practical and potential solution is to employ semi-supervised learning(SSL) to take advantage of the large amount of unlabeled data, which also contains the desired invariability across domains. 
The second limitation is that only one single modality image, the color fundus image, was involved in our experiments. Conducting experiments on other modality images would provide evidence for the effectiveness of our proposed method. In addition, involving other tasks, such as optic cup/disk and retinal vessel segmentation, would further extend our method for validating the causality in more applications. 
Thirdly, we only apply do-operation on the frequency space instead of the high-dimensional feature space generated by the network, which 
may be another suitable representation to disentangle $F_I$ and $F_V$ by carefully designing proxy tasks. Extending the current implementation of do-operation into feature space worth further investigating. Finally, during experiments, we ignore the class imbalance issue in each source domain to only focus on improving the average accuracy of five classes. Different class ratios in source and target domains may also affect the generalizable performance.

Future work would explore the incorporation of causality and SSL paradigm to involve more unlabeled data. The regular SSL setting assumes that each sample has an equal opportunity to be the unlabeled one, which may not be a valid assumption in some cases. For example, if considering the practical cost, the annotator usually tends to label the representative images, which contain more apparent key features related to the downstream task, such as the retina hemorrhage regions (the key metrics to grade DR) in our task. In this case, the representative ones have less chance of being the unlabeled set. Therefore, we intend to mitigate this selection bias through the causal perspective to encourage the model to learn invariant features across domains when utilizing unlabeled images in training. Furthermore, the causality would also be helpful in large visual models~\cite{qiu2023large,shi2023sam} recently emerging with massive parameters and datasets to potentially enhance in-context learning on huge datasets. In medical scenarios, leveraging causal correlations between medical concepts and clinical findings is critical and important for generalist medical artificial intelligence (GMAI)\cite{moor2023foundation} to carry out diverse tasks. Therefore, exploring causal learning in the mentioned settings will also be in our future research plans.


\section{Conclusions}\label{sec:conclusions}
In this work, we present a generalizable DR grading system by incorporating causality into the training pipeline to learn invariant features across multiple domains. Specifically, a structure causal model (SCM) is first proposed to model the fundus imaging process by analyzing the involved factors and their relationships. Then, we determine the features associated with spurious correlations and propose virtual interventions implemented by do-operation to cut off these correlations for better performance on images from unseen domains. To evaluate the performance, we collect four public fundus datasets associated with DR grading and reorganize them into 4 non-overlapping domains based on imaging devices to build a benchmark, i.e., 4DR. Finally, we conduct comprehensive experiments on this benchmark to demonstrate the effectiveness of our proposed CauDR framework. In the future, we will further extend this causality-inspired DG paradigm to more modalities and tasks.

\section*{Declaration of competing interest}
The authors declare that they have no conflict of interest.

{\small
\bibliographystyle{ieee_fullname}
\bibliography{reference}
}

\end{document}